\documentclass[sigconf]{acmart}

\AtBeginDocument{%
  \providecommand\BibTeX{{%
    \normalfont B\kern-0.5em{\scshape i\kern-0.25em b}\kern-0.8em\TeX}}}

\setcopyright{acmcopyright}
\copyrightyear{2018}
\acmYear{2018}
\acmDOI{10.1145/1122445.1122456}

\begin{document}

\title{A Multimodal Late Fusion Model for E-Commerce Product Classification}

\author{Ye Bi}
\affiliation{%
  \institution{Ping An Technology(Shenzhen)}
  \city{Shanghai}
  \country{China}
}
\email{yebi1313@163.com}

\author{Shuo Wang}
\affiliation{%
  \institution{Ping An Technology(Shenzhen)}
  \city{Shanghai}
  \country{China}}
\email{wangshuo240@pingan.com.cn}

\author{Zhongrui Fan}
\affiliation{%
  \institution{Ping An Technology(Shenzhen)}
  \city{Shanghai}
  \country{China}}
\email{fanzhongrui468@pingan.com.cn}

\renewcommand{\shortauthors}{Ye Bi, Shuo Wang and Zhongrui Fan, et al.}

\begin{abstract}
  The cataloging of product listings is a fundamental problem for most e-commerce platforms.
  Despite promising results obtained by unimodal-based methods, it can be expected that their performance can be further boosted by the consideration of multimodal product information.
  In this study, we investigated a multimodal late fusion approach based on text and image modalities to categorize e-commerce products on Rakuten. 
  Specifically, we developed modal specific state-of-the-art deep neural networks for each input modal, and then fused them at the decision level.
  Experimental results on Multimodal Product Classification Task of SIGIR 2020 E-Commerce Workshop Data Challenge\footnote{https://sigir-ecom.github.io/data-task.html} demonstrate the superiority and effectiveness of our proposed method compared with unimodal and other multimodal methods.
  Our team named \emph{pa\_curis} won the 1st place with a macro-F1 of 0.9144 on the final leaderboard\footnote{https://sigir-ecom.github.io/data-task.html\#scoreboard}.
\end{abstract}





\maketitle

\section{Introduction}
Product classification plays an important role in the e-commerce platform, with applications ranging from personalized search and recommendations to query understanding.
Categorizing products precisely helps e-commerce websites provide customers with a better shopping experience~\cite{DBLP:journals/corr/abs-1907-00420}. 
Manual annotation approach is not feasible for large-scale industrial deployment, so there remains a need to develop automatic product categorization systems.

However, one should note that the construction of such systems is also a challenging problem. 
The number of product categories could be enormous and the distribution of product quantities across categories could be highly unbalanced.
Moreover, there could be a large amount of noise in the textual and image data of products.
Massive efforts have been made to deal with the important yet tough problem, which can be broadly divided into two categories including unimodal based approaches and multimodal based approaches.
Particularly, the former either trains an image classifier based on product images or trains a text classifier based on textual information to categorize the product~\cite{cevahir-murakami-2016-large, DBLP:journals/corr/abs-1812-05774, xia-etal-2017-large}, whereas the latter attempts to build a classifier which combines product information from multiple modalities~\cite{DBLP:journals/corr/abs-1907-00420, DBLP:journals/corr/ZahavyMKM16, inproceedings}.
Most fusion techniques for multimodal learning can be grouped into feature-level fusion and decision-level fusion~\cite{DBLP:journals/corr/ZahavyMKM16}. 

The goal of SIGIR 2020 E-Commerce Workshop Data Challenge is to solve a fairly large-scale multimodal (text and image) product classification problem.
Given a training set of products with title, description and image information, and their corresponding product type codes, the participants need to predict the corresponding product type codes for an unseen held out test set of products.
To tackle this challenge, we explored the feature-level and the decision-level fusion scheme to leverage multimodal product information.
And the decision-level fusion scheme achieved better classification performance than the feature-level one under our experimental settings. 
For the decision-level fusion, the modal-specific classifiers are built from textual and image modal data respectively in the first stage.
Then the late fusion strategy is learned from the class probabilities predicted by each modal classifier in the second stage. 

The rest of the paper is organized as follows: Section 2 describes the challenge dataset. 
Our solution is introduced in Section 3 in details.
We show the experiments and results in the next Section 4. 
Finally, we conclude our analysis of the challenge.

\section{Dataset}
\begin{figure}[h]
  \centering
  \includegraphics[width=\linewidth]{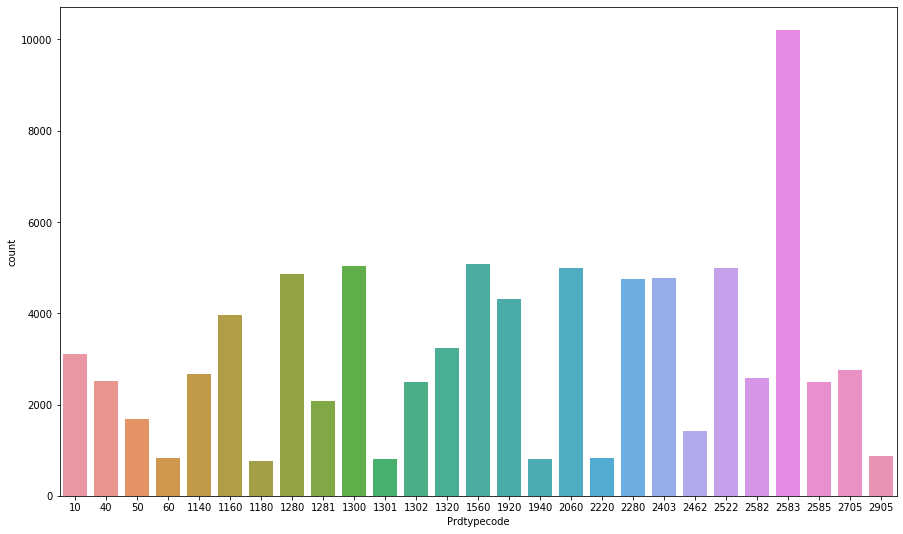}
  \caption{The frequency distribution of categories}
  \Description{The frequency distribution of categories}
  \label{class_dist}
\end{figure}

In this section, we give a brief introduction of the challenge dataset.
The organizer released approximately 99K product listings in tsv format, including 84,916 samples for training, 937 samples for phase 1 testing and 8435 samples for phase 2 testing. 
The dataset consists of product titles, descriptions, images and their corresponding product type codes.
There are 27 product categories in the training dataset and the number of product samples in each category ranges from 764 to 10,209.
The frequency distribution of categories in the training dataset is shown in Figure~\ref{class_dist}.

\section{Methodology}

\begin{figure}[h]
  \centering
  \includegraphics[width=\linewidth]{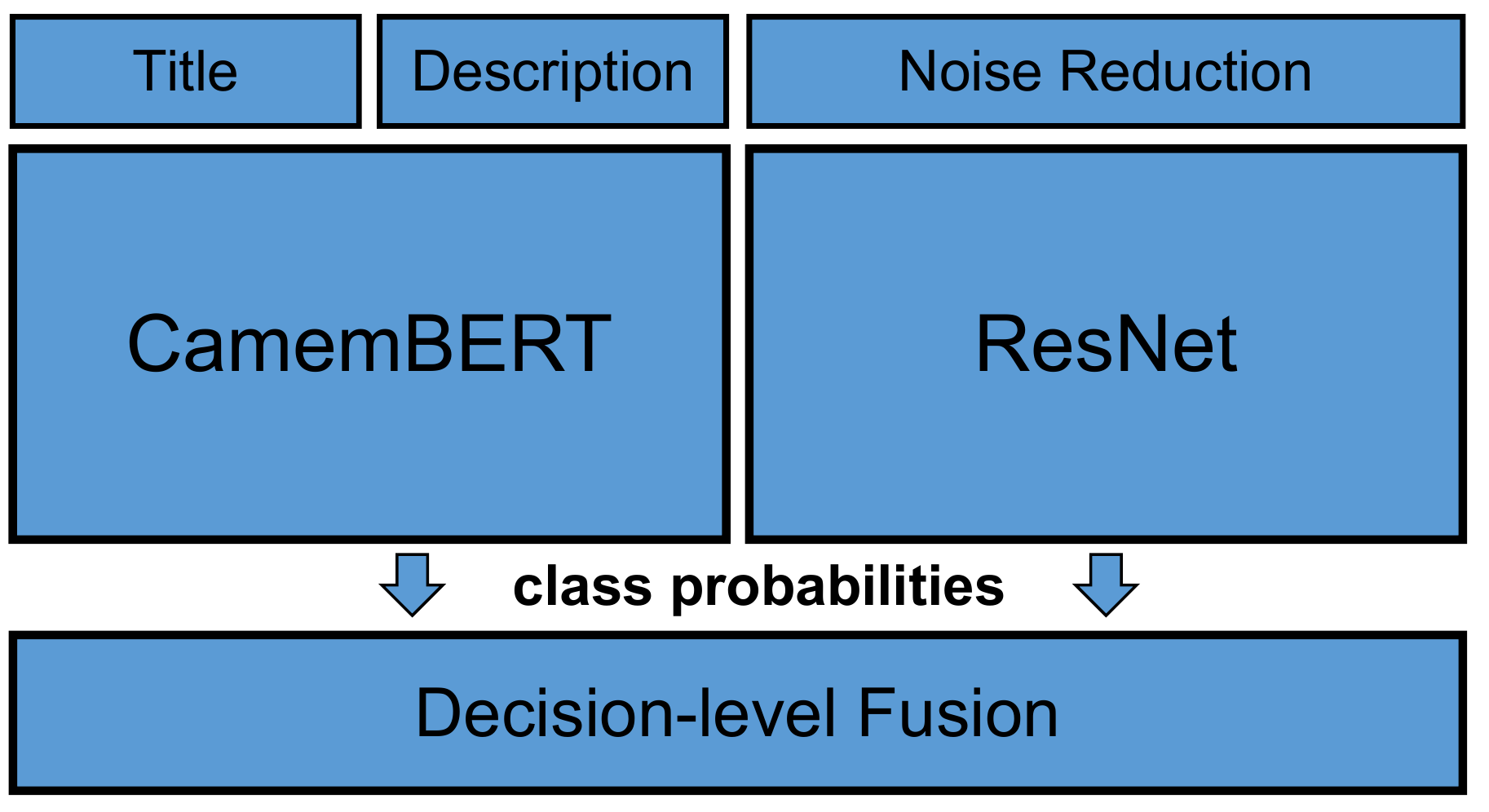}
  \caption{The model overview of multimodal product classification.}
  \Description{The model overview.}
  \label{overview}
\end{figure}

In this section, we introduce our proposed methods for the multimodal product classification task.
We first introduce the text and image based product classifier respectively and then we describe the fusion method in details.
The ensemble strategy is presented in the last. 
The overview of the proposed methods is shown in Figure~\ref{overview}.
The source code will be released on github~\footnote{https://github.com/Wang-Shuo/SIGIR2020Challenge} soon.

\subsection{Text classifier}
\label{subsec3.1}
The emergence of pretrained models~(PTMs) has brought natural language processing~(NLP) to a new era~\cite{Qiu2020PretrainedMF}. 
Recently, substantial work has shown that PTMs on the large corpus can learn universal language representations, which are beneficial for downstream NLP tasks and can avoid training a new model from scratch.
For the text based product classifier, we leveraged the state-of-the-art French PTM CamemBERT~\cite{Martin2020CamemBERTAT} since most of the product titles and descriptions are written in French.

\subsubsection{preprocessing}
Preprocessing is the preliminary step for most NLP tasks and responsible for the final performance to some extent.
We simply remove the excessive space and some HTML tags like \emph{L\&\#39} and \emph{<p>} from product title and description texts.

\subsubsection{text model}
CamemBERT is a state-of-the-art language model for French based on the RoBERTa~\cite{DBLP:journals/corr/abs-1907-11692} architecture pretrained on a large amount of French corpus, and achieves improved performance in many downstream tasks for French over previous monolingual and multilingual approaches. 
We concatenated the preprocessed product title and description text with [SEP] token and sent them as text pair into the CamemBERT model.
The pooled output of CamemBERT is taken as text representation, which is a 768 (CamemBERT-base) or 1024 (CamemBERT-large) dimensional vector.
A fully connected layer is employed as a linear classifier.
 
\subsection{Image classifier}
\label{subsec3.2}
Pretrained models are not just available for NLP tasks but also computer vision applications.
For the image based product classifier, we employed the commonly used ResNet152 network~\cite{DBLP:journals/corr/HeZRS15} pretrained on the ImageNet dataset.
The image classifier is trained using destruction and construction learning~(DCL)~\cite{Chen_2019_CVPR}, a fine-grained image recognition framework. 
Moreover, we applied the noise reduce techniques on the image dataset before training the image classifier.

\subsubsection{noise reduction}
\begin{figure}[h]
  \centering
  \includegraphics[width=\linewidth]{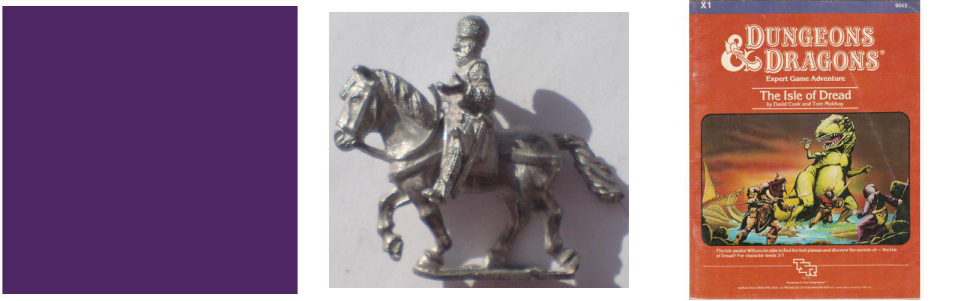}
  \caption{The sample images of \emph{Prdtypecode} 1180}
  \Description{The sample images of \emph{Prdtypecode} 1180.}
  \label{noisy_images}
\end{figure}

By looking at the pictures of each type of product in the training set, we found that there may be some label errors.
For example, the images shown in the Figure~\ref{noisy_images} are sampled from Prdtypecode 1180.
Although we don't know exactly what types of product the code 1180 refers to, it's clear that the image on the left should not fall into that category.
In other words, the left image could be noise.

If the classifier is trained with these noisy images directly, its performance could be degraded.
In view of this, we attempted to find label errors in the image dataset with an open source tool \emph{cleanlab}~\footnote{https://github.com/cgnorthcutt/cleanlab}, a framework powered by the theory of confident learning~\cite{northcutt2019confidentlearning}.
Specifically, we trained multiple ResNet50 image classifiers to compute the predicted product category probabilities for all the training samples in a cross-validation manner.
Then the cleanlab tool could utilize the matrix of predicted probabilities to find noisy samples, ordered by likelihood of being an error.
We removed the top ${10\%}$ noisy samples from the training set. 

\subsubsection{image model}
Although there are 27 product type codes in the training set, they belong to only 4 top level categories~(Child, Books, Household and Entertainment).
This implies that the differences of images among different product subcategories within the same top level category could be small.
So we treated the image based product classifier as a fine-grained image recognition task.
The DCL method has shown its effectiveness on the fine-grained product recognition task and won first place in multiple product recognition challenges.
Following the DCL scheme, we first fine-tuned the ResNet152 pretrained on ImageNet using ${224\times224}$ images, and then we reloaded the trained model and fine-tuned on higher resolution (${448\times448}$) with DCL configuration, to enhance the feature representation ability for backbone network.
The image augmentation techniques were performed during the training process, including image resize, random crop, rotation and horizontal flip.

\subsection{Fusion method}
To tackle the multimodal product classification task, we mainly tried two types of multimodal fusion methods which can be broadly divided into two categories~\cite{DBLP:journals/corr/ZahavyMKM16}: feature level fusion and decision level fusion.   
Regardless of the fusion strategy, we use the ResNet and CamemBERT network to extract features of the image and text, respectively.

\subsubsection{feature-level fusion}
Feature-level fusion method leverages multimodal information by concatenating the features extracted from specific unimodal network as the multimodal representation vector, followed by an additional classifier.
We tried different unification approaches such as concatenation, summation and even by means of attention mechanism. 
We also experimented with different training strategies such as end-to-end and step-by-step~\cite{DBLP:journals/corr/ZahavyMKM16}.
Despite all of these experiments, the best results that we achieved for feature-level fusion were inferior to those of the text unimodal classifier.
This leads us to turn to the decision-level fusion scheme. 

\subsubsection{decision-level fusion}
In this approach, an input-specific classifier is learned for each modality and then a fusion strategy is learned from the class probabilities predicted by each modal classifier.
Decision-level fusion has shown better performance than feature-level fusion in the product categorization task according to the study by Zahavy et al~\cite{DBLP:journals/corr/ZahavyMKM16}.
We employed this fusion scheme and trained the multimodal product classifier in a two-stage approach.
To be specific, in the first stage, we trained the text and image classifier described in ~\ref{subsec3.1} and ~\ref{subsec3.2}, respectively.
In the second stage, we designed a shallow neural network classifier that took all the class probabilities from the text and image classifier as input, and output the 27-class probabilities.

\subsection{Ensemble strategy} 
In the model ensemble stage, simple majority voting is used to ensemble the multiple classifiers.
Concretely, we ensembled 12 classifiers with the decision-level fusion approach, which were generated from different model configurations as follows:
\begin{itemize}
    \item different configurations for the text classifier, such as different backbone networks~(CamemBERT-base and CamemBERT-large), learning rates and batch size.
    \item different configurations for the image classifier. For example, whether to use the clean dataset after denoising and whether to use the DCL training method could produce different candidate models. Moreover, models saved at the late training phase were also exploited. 
    \item different configurations for the decision-level fusion. We tried 1-layer and 2-layer neural network classifiers.
\end{itemize}

\section{Experiment}
In this Section, we provide experimental settings and results.

\subsection{Experimental Settings}
We divided the full 84916 labeled samples randomly into training and validation set at a ratio of 9:1.
For the noise reduction part, we performed 4-fold cross-validation to compute the matrix of predicted probabilities.
There are about 2120 training samples in the top ${10\%}$ noisy samples given by the cleanlab.
The text classifier is trained with \emph{AdamW} optimizer and the initial learning rate is 3e-5 or 5e-5 and decreases linearly after a warmup period.
The batch size is set to 64 or 128 and the number of epochs is set 40. 
The image classifier is trained with \emph{SGD} optimizer and the initial learning rate is 0.01 and decreases every 12 epochs at a rate of 0.1.
The backbone network is trained for 60 epochs in the first stage and then fine-tuned with the DCL method for 20 epochs in the second stage.
We adopted the code~\footnote{https://github.com/JDAI-CV/DCL} released by the author.
For the decision-level fusion scheme, we trained the decision-level fusion policy on the validation set in a 8-fold cross-validation manner.
The hidden size in the 2-layer fusion neural network classifier is set to 6.
The fusion policy neural networks were trained with \emph{Adam} optimizer and the learning rate is set to 0.01.
We trained the policy network for 40 epochs and kept the checkpoint that has the highest macro-f1 score on the validation set.
The predicted results of 8 trained policy networks are ensembled by majority voting when inferencing on the test set.  

\subsection{Results}

Here we compared the performance of our method with different settings.
The results on the test set are shown in Table~\ref{tab:results}.
The online results for some methods are not given due to the limit on the number of times the results could be submitted online.
From the results, we can see that the decision-level fusion scheme achieves better results than unimodal method and feature-level fusion method.
And the ensemble strategy also brings about performance improvement on the online results. 

\begin{table}
  \caption{The online Macro-F1 results(\%)}
  \label{tab:results}
  \begin{tabular}{ccl}
    \toprule
    Method & Phase 1 & Phase 2\\
    \midrule
    Uni-Image Classifier&  69.21 & -\\
    Uni-Text Classifier& 89.93 & -\\
    Feature-level Fusion& 89.87 & -\\
    Decision-level Fusion& 90.94 & 90.17\\
    Ensemble & - & 91.44\\
  \bottomrule
\end{tabular}
\end{table}

\section{Conclusion}

In this paper, we introduced our solutions for the multimodal product classification task in details on the SIGIR 2020 E-Commerce Workshop Data Challenge. 
Extensive experiments were conducted on the challenge dataset and the results proved the effectiveness of our method.
In the future work, we will perform some more experiments and get the results of some methods on the full test set as soon as the full test set is released with ground truth labels.

\bibliographystyle{ACM-Reference-Format}
\bibliography{sigir2020}










\end{document}